\PassOptionsToPackage{unicode}{hyperref}
\PassOptionsToPackage{hyphens}{url}
\PassOptionsToPackage{dvipsnames,svgnames,x11names}{xcolor}
\documentclass[
  11pt,
  letterpaper,
]{article}
\usepackage{xcolor}
\usepackage[margin=1in]{geometry}
\usepackage{amsmath,amssymb}
\usepackage{iftex}
\ifPDFTeX
  \usepackage[T1]{fontenc}
  \usepackage[utf8]{inputenc}
  \usepackage{textcomp} 
\else 
  \usepackage{unicode-math} 
  \defaultfontfeatures{Scale=MatchLowercase}
  \defaultfontfeatures[\rmfamily]{Ligatures=TeX,Scale=1}
\fi
\usepackage{lmodern}
\ifPDFTeX\else
\fi
\IfFileExists{upquote.sty}{\usepackage{upquote}}{}
\IfFileExists{microtype.sty}{
  \usepackage[]{microtype}
  \UseMicrotypeSet[protrusion]{basicmath} 
}{}
\makeatletter
\@ifundefined{KOMAClassName}{
  \IfFileExists{parskip.sty}{%
    \usepackage{parskip}
  }{
    \setlength{\parindent}{0pt}
    \setlength{\parskip}{6pt plus 2pt minus 1pt}}
}{
  \KOMAoptions{parskip=half}}
\makeatother
\usepackage{longtable,booktabs,array}
\usepackage{caption}
\usepackage{calc} 
\usepackage{etoolbox}
\makeatletter
\patchcmd\longtable{\par}{\if@noskipsec\mbox{}\fi\par}{}{}
\makeatother
\global\tracinglostchars=3
\usepackage{xurl}
\usepackage{seqsplit}
\usepackage{etoolbox}
\usepackage{caption}
\usepackage{float}
\usepackage{needspace}
\AtBeginEnvironment{longtable}{\small}
\usepackage{bookmark}
\IfFileExists{xurl.sty}{\usepackage{xurl}}{} 
\makeatletter
\@ifundefined{xmpquote}{}{}
\makeatother
\hypersetup{
  pdftitle={NCP: Verifier-Guided Optimization of Weight-Space Control Policies},
  pdfauthor={Ekrem Aksoy},
  colorlinks=true,
  linkcolor={blue},
  filecolor={Maroon},
  citecolor={Blue},
  urlcolor={blue},
  pdfcreator={LaTeX via pandoc}}

\title{NCP: Verifier-Guided Optimization of Weight-Space Control Policies}
\usepackage{etoolbox}
\makeatletter
\providecommand{\subtitle}[1]{
  \apptocmd{\@title}{\par {\large #1 \par}}{}{}
}
\makeatother
\subtitle{Toward Recursive Self-Improvement through Neural Flow Rules}
\author{Ekrem Aksoy}
\date{November 5, 2026}

\begin{document}
\maketitle
\begin{abstract}
We present the Neural Control Plane (NCP), a closed-loop system that
observes language-model execution, synthesizes candidate weight-space
intervention rules, evaluates them, and maintains a versioned table of
admitted policies. NCP separates a pretrained model's inference data
plane from the control policy that selects temporary parameter
interventions. W-ISA supplies transactional execution and parameter
restoration; NCP supplies telemetry, candidate synthesis, admission
criteria, priorities, and rule lifetime management. The optimization
state consists of active rules and an archive of evaluated candidates,
while base-model parameters are restored after each request. In a local
SmolLM-135M-Instruct experiment, NCP generated three candidates from
development telemetry and evaluated them on separate task examples and a
4,095-target WikiText-2 slice. One projection reduced validation token
repetition from 0.4133 to 0.2191, with a perplexity change of -0.0574\%.
All candidates produced zero newly correct exact answers, so the
configured admission gate archived all three and retained the original
active policy. Fresh-process repetition and saved-table resumption
reproduced the measured outcomes. NCP provides an executable framework
for studying recursive self-improvement at the control-policy level
through explicit synthesis, verification, and state transitions.
\end{abstract}

\textbf{Keywords:} neural control plane; recursive self-improvement;
inference-time optimization; weight-space interventions; model editing;
transactional execution.

\section{Introduction}\label{introduction}

Inference-time adaptation can be expressed through changes to prompts,
intermediate activations, auxiliary memory, or model parameters. Each
choice creates a different optimization state and execution lifecycle.
NCP studies the policy that decides which weight-space intervention to
apply to an incoming request.

We explicitly credit Rikiya Takehi, Jayden Teoh, Alex Zhang, and Omar
Khattab's \emph{Sparks of Weight-Space Programming: LLMs can natively
read \& write neural network weights} {[}11{]} as a conceptual
foundation. Their November 2, 2026 article presents reasoning-driven
weight inspection and editing, describes a tool-assisted loop-reduction
experiment on Qwen3.5-0.8B, and identifies recursive self-improvement as
a research direction. NCP develops a verifier-guided control-policy
architecture around this weight-space programming perspective, using
W-ISA for scoped execution.

The system observes task outcomes and output-token telemetry, extracts
an activation direction associated with development failures, and
constructs explicit match-action candidates. A verifier compares the
current policy with the policy that would result from admitting each
candidate. Accepted rules become part of the active flow table;
evaluation records preserve the outcomes of every trial.

Our architectural starting point is the separation of control and data
planes used in software-defined networking. OpenFlow provides a
flow-table interface through which a controller manages
packet-processing behavior {[}1{]}. NCP adopts this separation as a
software design: the language model performs tensor computation, the
W-ISA runtime executes parameter interventions, and the controller
selects and evaluates rules.

The recursive self-improvement (RSI) research objective is improvement
of this control policy across successive feedback cycles. An admitted
rule changes the behavior observed in later requests, which changes
subsequent telemetry and candidate generation. The implemented loop uses
a deterministic synthesizer and an explicit verifier, making each
proposed policy transition inspectable.

The central question is:

\begin{quote}
Can a language-model control policy be improved through repeated
generation, temporary execution, and verification of
activation-conditioned weight-space rules while preserving a stable base
parameter state?
\end{quote}

The present contribution is an executable architecture for this
question: a versioned neural flow table, deterministic candidate
generation, a routed-task and forced-action admission gate, and a
reproducible local evaluation protocol.

\section{Background and Related Work}\label{background-and-related-work}

\subsection{Weight-space programming}\label{weight-space-programming}

Takehi et al.~{[}11{]} study general-purpose LLMs reading and writing
weights under natural-language objectives, including experiments with
controlled forward-pass budgets. Their discussion of model correction
and recursive self-improvement motivates the NCP research objective. The
present implementation supplies deterministic candidate templates,
routed task evaluation, forced-action corpus checks, and evidence-bound
policy admission.

\subsection{Control-plane separation}\label{control-plane-separation}

OpenFlow exposes switch flow tables through an interface for adding and
removing entries {[}1{]}. The corresponding NCP objects are activation
predicates, W-ISA action programs, priorities, and rule lifetimes. The
analogy concerns programmatic control and policy management; NCP's
matching signals are neural activations, and its actions modify selected
parameter tensors within an inference scope.

\subsection{Feedback-driven
improvement}\label{feedback-driven-improvement}

Self-Refine iteratively revises outputs using model-generated feedback
{[}2{]}. Reflexion maintains linguistic feedback in episodic memory to
influence later trials {[}3{]}. NCP instead makes a table of executable
intervention rules the persistent optimization state. Its current
synthesis step is a deterministic Python procedure driven by observed
task labels and captured activations.

OPRO generates and evaluates candidate solutions using a language model
conditioned on earlier solutions and their scores {[}12{]}. DSPy
separates declarative language-model programs from metric-driven
pipeline optimization {[}13{]}. These studies, also referenced by Takehi
et al., provide context for separating candidate representation,
proposal generation, and evaluation. NCP applies this separation to a
table of activation-conditioned weight interventions.

\subsection{Neural interventions}\label{neural-interventions}

ROME and MEMIT construct parameter updates for factual associations
{[}4, 5{]}. Contrastive Activation Addition derives behavioral
directions from activation differences {[}6{]}. pyvene provides
configurable intervention schemes {[}7{]}, while NNsight supports
tracing, persistent intervention definitions, and conditional execution
{[}8{]}. These works provide related edit constructions and intervention
infrastructures.

NCP builds on the W-ISA execution interface described in the companion
manuscript {[}9{]}. W-ISA handles instruction validation, weight writes,
snapshots, synchronization, and restoration. NCP handles the question of
which program to propose, which evidence to collect, and which rule to
admit.

\subsection{Repetition and directional
interventions}\label{repetition-and-directional-interventions}

Pipis et al.~{[}14{]} analyze looping in reasoning models, including its
relationship to decoding temperature and temporally correlated errors.
Their study motivates measuring repetition separately from answer
correctness. NCP records token n-gram repetition, decoding-score
entropy, and exact-answer outcomes as distinct signals. Arditi et
al.~{[}15{]} provide a related example of studying model behavior
through a residual-stream direction; NCP's current projection candidates
use development-derived directions in a selected MLP input space.

\section{System Model}\label{system-model}

Let \(f_\theta\) denote a pretrained model with registered base
parameter state \(\theta\). Let \(\Phi_t\) be the active rule table at
controller cycle \(t\), and \(A_t\) the archive of evaluated candidates.
A request is routed through \(\Phi_t\) to choose a program \(P\) or the
identity program.

The system's persistent optimization state is

\[\mathcal{S}_t=(\Phi_t,A_t,e_t),\]

where \(e_t\) is the active-table version. The base parameter state
remains the reference state between requests.

The control recurrence is

\[\mathcal{T}_t=\operatorname{Observe}(f_\theta,\Phi_t,D_{\mathrm{dev}}),\]

\[\mathcal{C}_t=\operatorname{Synthesize}(\mathcal{T}_t),\]

\[(\Phi_{t+1},A_{t+1})=
\operatorname{Admit}\bigl(\Phi_t,A_t,\mathcal{C}_t,D_{\mathrm{val}},C_{\mathrm{gate}}\bigr).\]

Candidates are evaluated sequentially. Admission of an earlier candidate
changes the policy context used to evaluate a later candidate. Repeated
candidate identifiers retrieve their recorded decisions. A cycle with no
new candidate evaluations terminates the current search.

This recurrence defines the control-policy layer of the RSI research
program. Its objects are rule proposals, verification evidence, and
active-policy transitions. The experiment in Section 7 records a cycle
in which the archive grows and the active policy remains unchanged.

\section{Neural Flow Table}\label{neural-flow-table}

\subsection{Rule definition}\label{rule-definition}

A rule is represented as

\[r = \bigl(\mathrm{id},\,M,\,\ell,\,v,\,\tau,\,P,\,\pi,\,H,\,I,\,B\bigr),\]

where \(M\) identifies the base model; \(\ell\) specifies the activation
module, input/output location, and pooling method; \(v\) is a match
direction; \(\tau\) is a threshold; \(P\) is a W-ISA program; \(\pi\) is
priority; \(H\) and \(I\) are hard and idle timeouts; and \(B\) is an
optional generation-token budget.

Model identity includes the model identifier, checkpoint revision, and a
SHA-256 digest of parameter names, shapes, dtypes, and bytes. A rule
fingerprint covers its definition and numeric operands. The table owns
copies of submitted data.

For a nonzero captured activation \(h_\theta^\ell(x)\) and nonzero
direction \(v\), matching uses

\[s_r(x)=\frac{h_\theta^\ell(x)^Tv}{\lVert h_\theta^\ell(x)\rVert_2\lVert v\rVert_2},
\qquad s_r(x)>\tau.\]

The implementation assigns score zero to a zero activation. Higher
numeric priority is considered first, and equal-priority rules retain
registration order. The first matching rule is selected. Activations are
measured before the selected parameter intervention is applied.

\subsection{Lifecycle and evidence}\label{lifecycle-and-evidence}

The table records candidate, active, rejected, expired, and retired
states. Promotion requires accepted evidence for the rule's fingerprint
and the current active-table version. Changes to the active policy
invalidate evidence measured against an earlier version.

Accepted rule definitions persist in the table. Their weight effects
remain temporary and are reconstructed during matching requests. The
runtime holds a shared model lock, applies the selected program,
performs generation using a fresh KV cache, and restores the
request-entry weights.

Hard timeout is elapsed wall-clock time since promotion. Idle timeout is
elapsed time since promotion or the most recent completed selected
request. Expiration is applied during table inspection for dispatch or
admission; an in-flight request completes under its selected scope.
Persisted timestamps preserve elapsed time across process restart. Token
budgets are separate from these time-based fields.

Table snapshots preserve status, order, timestamps, hit counts,
definitions, and evidence. Persistence writes a flushed temporary JSON
file and atomically replaces the destination.

\section{Telemetry and Candidate
Synthesis}\label{telemetry-and-candidate-synthesis}

\subsection{Output and task
measurements}\label{output-and-task-measurements}

Each observation records generated text, token IDs, elapsed time,
token-budget completion, and per-step entropy of the decoding-score
distribution:

\[H_t=-\sum_j p_{t,j}\log p_{t,j}.\]

Repetition telemetry uses token n-grams. For \(G\) observed n-grams and
\(U\) distinct n-grams, the repeated fraction is

\[R=\frac{G-U}{G},\]

with \(R=0\) when \(G=0\). The implementation also records the first
adjacent repeated-block position and the maximum consecutive repetition
run. The default n-gram size is four. Terminal EOS is removed from the
repetition sequence.

Task scoring uses an explicit expected-answer specification. In the
recorded fixture, the requested format ends with
\texttt{Final\ answer:\ \textless{}number\textgreater{}}. The scorer
takes the last answer-pattern capture, or the complete response when the
pattern is absent, and applies case and trailing-period normalization
before exact comparison.

\subsection{Activation profiles}\label{activation-profiles}

Development outcomes label prompt-end activations. When both classes are
present, the profile is

\[v_{\mathrm{error}} = \mu_{\mathrm{failure}} - \mu_{\mathrm{success}}.\]

For an all-failure development set, the implementation uses
\(v_{\mathrm{error}}=\mu_{\mathrm{failure}}\) and records the source as
\texttt{failure\_centroid}. Capture and matching use the same activation
site and pooling method. Zero-norm directions produce an empty candidate
set for that site.

\subsection{Deterministic action
templates}\label{deterministic-action-templates}

For each configured transformer block and attention head, the initial
synthesizer constructs:

\Needspace{10.5\baselineskip}

{\def\LTcaptype{none} 
\begin{longtable}[]{@{}
  >{\raggedright\arraybackslash}p{(\linewidth - 2\tabcolsep) * \real{0.5000}}
  >{\raggedright\arraybackslash}p{(\linewidth - 2\tabcolsep) * \real{0.5000}}@{}}
\toprule\noalign{}
\begin{minipage}[b]{\linewidth}\raggedright
Candidate
\end{minipage} & \begin{minipage}[b]{\linewidth}\raggedright
W-ISA action
\end{minipage} \\
\midrule\noalign{}
\endhead
\bottomrule\noalign{}
\endlastfoot
MLP projection & Remove the profiled input direction from the block's
down-projection \\
Head scaling & Scale the selected head's columns in the attention output
projection \\
Combined & Execute head scaling followed by the MLP projection \\
\end{longtable}
}

The block and head dimensions are read from the loaded model. Candidate
identifiers are deterministically derived from the action and matching
configuration. Compilation checks the generated instruction sequence
against the actual parameters before evaluation.

\subsection{Controller-agnostic candidate
representation}\label{controller-agnostic-candidate-representation}

The \texttt{NeuralRule} representation separates a candidate's origin
from its execution and evaluation. It holds model identity, an
activation predicate, a W-ISA source program, numeric operands and
lifecycle fields. \texttt{TorchBackend.compile()} binds that source to
the model through the W-ISA compiler. Caller-constructed rules can
therefore use the supported instruction set independently of the
templates in Section 5.3.

The public path is \texttt{FlowTable.submit(rule)}, followed by
\texttt{\seqsplit{Verifier.evaluate(rule, table, cases, corpus\_tokens)}},
and then promotion or rejection using the returned evidence. The
verifier evaluates the proposed policy and the forced action; promotion
checks the accepted flag, fingerprint and active-table version. These
interfaces operate on trusted local rule and evidence data. The
controller's \texttt{step()} method assembles this path around the
deterministic synthesizer used in the recorded experiment.

\section{Verification and Policy
Admission}\label{verification-and-policy-admission}

\subsection{Routed task comparison}\label{routed-task-comparison}

The verifier evaluates the existing active table and the prospective
table containing the candidate. Both runs use the same validation cases
and the table's actual priority order.

Let \(c\) be the number of incorrect-to-correct transitions, \(r\) the
number of correct-to-incorrect transitions, \(m\) the number of
validation cases that selected the candidate, and

\[\Delta R=\bar R_{\mathrm{before}}-\bar R_{\mathrm{after}}.\]

The initial policy requires \(c\ge1\), \(r=0\), \(m>0\), and
\(\Delta R\ge0\). Thresholds are caller-configurable. The gate also
requires progress in task correctness or repetition: \(c>r\) or
\(\Delta R>0\).

\subsection{Forced-action corpus
comparison}\label{forced-action-corpus-comparison}

The candidate program is forced active across the entire verifier corpus
slice, independently of the trigger. This measurement evaluates the
action under a common exposure protocol, while routed task evaluation
measures the policy that would be deployed.

For mean token NLL \(\bar\ell_0\) under original weights and
\(\bar\ell_P\) under the forced action, the relative perplexity change
is

\[\Delta\mathrm{PPL}_{\%}=100\left[\exp(\bar\ell_P-\bar\ell_0)-1\right].\]

The recorded policy sets a maximum increase of 0.5\%. Original-corpus
scores are cached by token digest and scoring protocol.

\subsection{Spectral and restoration
checks}\label{spectral-and-restoration-checks}

For each edited parameter, the verifier records \(\lVert W\rVert_2\),
\(\lVert\Delta W\rVert_2\), \(\lVert\Delta W\rVert_F\), and the relative
spectral change. A configured relative spectral budget adds a
corresponding admission condition.

Model parameter digests are checked around verification. The original
weights are restored after each routed request and forced-action trial.
Before promotion, evidence is checked against the candidate fingerprint
and active-table version.

Under the recorded defaults, admission can be written as

\[\begin{aligned}
\operatorname{Accept}(P) ={}& [m > 0] \land [c \ge 1] \land [r = 0] \\
& {}\land [\Delta R \ge 0] \land [\Delta\mathrm{PPL}_{\%} \le 0.5] \land E,
\end{aligned}\]

where \(E\) denotes exact parameter restoration and valid admission
provenance. Configured spectral bounds are conjoined with this
expression. Accepted candidates become active rules; other evaluated
candidates retain their measurements and decision reasons in the
archive.

\section{Architectural
Microbenchmark}\label{architectural-microbenchmark}

\subsection{Protocol}\label{protocol}

The recorded run uses
\texttt{\seqsplit{HuggingFaceTB/SmolLM-135M-Instruct}}, revision
\texttt{\seqsplit{b5f4b394abe60fe0e4d666139e57bfce520302a9}}, with FP32
weights on MPS and CPU FP64 intervention arithmetic. The runtime is
PyTorch 2.15.0 with Python 3.12.9. Generation is greedy with a 128-token
budget.

The nine-question handwritten fixture is an architectural microbenchmark
of telemetry, candidate generation, admission and persistence. It
contains three development, three validation, and three final-test
questions with disjoint IDs and prompts. Development questions cover
addition, sequence continuation, and counting grouped objects.
Validation and final-test questions use separate arithmetic scenarios.
Final-test inference begins after candidate selection finishes.

The language-modeling corpus is the cached WikiText-2 raw test split
{[}10{]}. Its tokenized stream is partitioned into two disjoint
4,096-token slices. The first supplies the verifier corpus and the
second supplies final corpus evaluation. Each slice scores 4,095 targets
once, with context length 512 and stride 256. These are explicitly
bounded corpus slices.

\subsection{Generated candidates}\label{generated-candidates}

Development exact-answer scoring was 0/3. Synthesis therefore used the
failure centroid at the last prompt token's input to
\texttt{\seqsplit{model.layers.9.mlp.down\_proj}}. This direction has
1,536 dimensions and norm 3.6363433948.

The three candidates target the tenth transformer block: an MLP
projection, scaling query head 0's output contribution by zero, and
their ordered combination. The cosine threshold is 0.8. Each candidate
was selected on all three validation prompts.

\subsection{Task and corpus results}\label{task-and-corpus-results}

Baseline validation repetition fraction was 0.41333333. Baseline
verifier perplexity was 23.74291741.

\Needspace{10.5\baselineskip}

{\def\LTcaptype{none} 
\begin{longtable}[]{@{}
  >{\raggedright\arraybackslash}p{(\linewidth - 10\tabcolsep) * \real{0.2000}}
  >{\raggedleft\arraybackslash}p{(\linewidth - 10\tabcolsep) * \real{0.1900}}
  >{\raggedleft\arraybackslash}p{(\linewidth - 10\tabcolsep) * \real{0.1400}}
  >{\raggedleft\arraybackslash}p{(\linewidth - 10\tabcolsep) * \real{0.2000}}
  >{\raggedleft\arraybackslash}p{(\linewidth - 10\tabcolsep) * \real{0.1600}}
  >{\raggedright\arraybackslash}p{(\linewidth - 10\tabcolsep) * \real{0.1100}}@{}}
\toprule\noalign{}
\begin{minipage}[b]{\linewidth}\raggedright
Candidate
\end{minipage} & \begin{minipage}[b]{\linewidth}\raggedleft
Validation repetition fraction
\end{minipage} & \begin{minipage}[b]{\linewidth}\raggedleft
Newly correct answers
\end{minipage} & \begin{minipage}[b]{\linewidth}\raggedleft
Forced-action perplexity
\end{minipage} & \begin{minipage}[b]{\linewidth}\raggedleft
PPL change
\end{minipage} & \begin{minipage}[b]{\linewidth}\raggedright
Decision
\end{minipage} \\
\midrule\noalign{}
\endhead
\bottomrule\noalign{}
\endlastfoot
MLP projection & 0.21906173 & 0 & 23.72929551 & -0.0573725\% &
Archived \\
Head scale 0.0 & 0.36800000 & 0 & 23.78039795 & +0.1578599\% &
Archived \\
Combined & 0.51733333 & 0 & 23.77012175 & +0.1145787\% & Archived \\
\end{longtable}
}

All three candidates met the configured perplexity budget. The
corrected-answer threshold was one; the active table consequently
remained at zero rules. The projection and head-scaling candidates
reduced repetition, while their combination increased it under this
fixture.

The MLP projection had relative spectral change 0.17329298; head scaling
had 0.46918526. The combined program recorded both parameter changes
separately. Every trial restored the base parameter digest.

Final exact-answer scoring was 0/3 for both original and routed
execution. Both recorded a repetition fraction of 0.256. Final-slice
baseline perplexity was 33.72129164. With the active policy unchanged,
the final outputs matched the baseline policy.

\subsection{Repeat, resume, and test
coverage}\label{repeat-resume-and-test-coverage}

The initial end-to-end run took 44.38 seconds. A new-process two-cycle
run took 48.37 seconds and reproduced first-cycle generated outputs,
token entropies, candidate fingerprints, spectral measurements, and
perplexity values. Its second cycle recognized the three recorded
candidate identifiers. A saved-table resume in a third process took
15.56 seconds and preserved the decisions and final outputs.

Independent reconstruction from the saved per-window NLL sums confirmed
all 4,095 target counts and aggregate perplexity values. Direct
reconstruction of corrected-case counts agreed with the admission
records.

The NCP suite contains 25 passing tests. Controlled fixtures exercise
both promotion and rejection through actual W-ISA parameter writes.
Other cases cover hard and idle expiration, elapsed TTL across restart,
stale evidence, priority shadowing, definition copying, token-budget
dispatch, forced-action PPL for unmatched inputs, cancellation, and
scalar-parameter fingerprints.

The recorded SmolLM run demonstrates the complete candidate-to-decision
workflow and archive persistence. Its active-policy transition is
\(\Phi_{t+1}=\Phi_t=\varnothing\), with three evaluated candidates added
to the archive. The controlled promotion fixtures exercise the
complementary active-table transition.

\section{Control-Policy Optimization}\label{control-policy-optimization}

The optimization state has two parts with distinct effects. The active
table changes inference routing, while the archive records search
history. In the measured cycle the archive grew by three candidates and
the active table stayed empty. On the next cycle, deterministic
candidate identifiers retrieved the existing decisions. Saved-table
resumption retained the same search state across process boundaries.

The verifier separates task progress from repetition and corpus cost. In
the projection trial, repetition decreased by 0.19427160 and the
forced-action corpus score satisfied the configured budget. The
corrected-answer count remained zero, so the rule was archived under the
one-correction admission criterion. This decision illustrates how the
configured objective determines a policy transition rather than treating
every change in an auxiliary signal as an improvement.

Proposal construction and acceptance remain separate operations. The
representation in Section 5.4 carries a program regardless of its
producer, while the verifier binds evidence to the exact candidate and
current policy. Once a rule is active, priority matching and expiry
determine its execution lifetime; the W-ISA scope determines the
lifetime of its parameter effects.

The companion W-ISA study {[}9{]} measures this execution layer on
SmolLM-135M and Qwen2.5-1.5B, including a 45-request comparison between
materialized rank-one updates and equivalent PyTorch hooks. NCP's
microbenchmark measures the controller lifecycle on SmolLM. The shared
abstraction is an explicit intervention program whose temporary weight
effects are restored before the next policy evaluation.

\section{Reproducibility and
Implementation}\label{reproducibility-and-implementation}

The implementation contains the schema, flow table, telemetry, local
backend, candidate synthesizer, verifier, and controller. It reuses
W-ISA's instruction compiler and request-scoped runtime. Public APIs
include \texttt{NeuralControlPlane}, \texttt{NeuralRule},
\texttt{FlowTable}, \texttt{TorchBackend}, \texttt{TaskCase},
\texttt{GatePolicy}, and \texttt{Verifier}.

Rule tables are trusted local control data. Model identity, rule
fingerprints, active-table versions, corpus-token hashes, and task
fixtures are retained in the reports. Persistence uses a flushed
temporary file followed by atomic replacement. Runtime scopes own
parameter restoration; table promotion persists policy definitions.

Separate records retain the initial run, fresh-process repeat and
saved-table resumption. The reasoning fixture defines expected answers
and disjoint task splits; the tests exercise scoring and lifecycle
transitions. The companion W-ISA manuscript {[}9{]} describes the
execution substrate separately.

The experiment runner exposes options for the device, task fixture,
corpus, candidate layers and heads, thresholds, timeouts, token budgets,
and output location. Section 7 specifies the settings used for the
recorded microbenchmark.

Code uses the MIT License, copyright 2026 Ekrem Aksoy. Model and corpus
materials retain their upstream attribution and license terms.

\section{Conclusion}\label{conclusion}

NCP makes a language model's intervention policy an explicit
optimization state. Telemetry generates candidates; W-ISA executes them
within owned scopes; a verifier evaluates routed task effects and
forced-action corpus effects; and a versioned flow table records
admission decisions. The initial local experiment produced three
reproducible candidate evaluations, retained the configured correctness
criterion, and restored the base model after every trial. This
architecture provides a concrete basis for studying recursive
self-improvement through successive, evidence-bound changes to neural
flow policies.

\section*{References}\label{references}
\addcontentsline{toc}{section}{References}

\small\raggedright\setlength{\parskip}{3pt}

{[}1{]} Nick McKeown, Tom Anderson, Hari Balakrishnan, Guru Parulkar,
Larry Peterson, Jennifer Rexford, Scott Shenker, and Jonathan Turner.
\emph{OpenFlow: Enabling Innovation in Campus Networks}. ACM SIGCOMM
Computer Communication Review, 38(2):69--74, 2008.
\url{https://doi.org/10.1145/1355734.1355746}

{[}2{]} Aman Madaan et al.~\emph{Self-Refine: Iterative Refinement with
Self-Feedback}. arXiv:2303.17651, 2023.
\url{https://arxiv.org/abs/2303.17651}

{[}3{]} Noah Shinn et al.~\emph{Reflexion: Language Agents with Verbal
Reinforcement Learning}. arXiv:2303.11366, 2023.
\url{https://arxiv.org/abs/2303.11366}

{[}4{]} Kevin Meng, David Bau, Alex Andonian, and Yonatan Belinkov.
\emph{Locating and Editing Factual Associations in GPT}. NeurIPS, 2022.
\url{https://arxiv.org/abs/2202.05262}

{[}5{]} Kevin Meng, Arnab Sen Sharma, Alex Andonian, Yonatan Belinkov,
and David Bau. \emph{Mass-Editing Memory in a Transformer}.
arXiv:2210.07229. \url{https://arxiv.org/abs/2210.07229}

{[}6{]} Nina Panickssery et al.~\emph{Steering Llama 2 via Contrastive
Activation Addition}. arXiv:2312.06681.
\url{https://arxiv.org/abs/2312.06681}

{[}7{]} Zhengxuan Wu et al.~\emph{pyvene: A Library for Understanding
and Improving PyTorch Models via Interventions}. arXiv:2403.07809.
\url{https://arxiv.org/abs/2403.07809}

{[}8{]} NNsight documentation. \emph{Model Editing} and
\emph{Conditional Interventions}. Accessed November 3, 2026.
\url{https://nnsight.net/features/7_model_editing/} ;
\url{https://nnsight.net/vllm/conditional/}

{[}9{]} Ekrem Aksoy. \emph{W-ISA: Transactional Execution of
Weight-Space Intervention Programs}. Companion manuscript, 2026.

{[}10{]} Stephen Merity, Caiming Xiong, James Bradbury, and Richard
Socher. \emph{Pointer Sentinel Mixture Models}. arXiv:1609.07843, 2016.
\url{https://arxiv.org/abs/1609.07843}. Dataset distribution:
\url{https://huggingface.co/datasets/Salesforce/wikitext}

{[}11{]} Rikiya Takehi, Jayden Teoh, Alex Zhang, and Omar Khattab.
\emph{Sparks of Weight-Space Programming: LLMs can natively read \&
write neural network weights}. Blog post, November 2, 2026.
\url{https://rikiyat.github.io/blog/weight-space-programming/}

{[}12{]} Chengrun Yang, Xuezhi Wang, Yifeng Lu, Hanxiao Liu, Quoc V. Le,
Denny Zhou, and Xinyun Chen. \emph{Large Language Models as Optimizers}.
ICLR, 2024. \url{https://arxiv.org/abs/2309.03409}

{[}13{]} Omar Khattab et al.~\emph{DSPy: Compiling Declarative Language
Model Calls into Self-Improving Pipelines}. ICLR, 2024.
\url{https://arxiv.org/abs/2310.03714}

{[}14{]} Charilaos Pipis, Shivam Garg, Vasilis Kontonis, Vaishnavi
Shrivastava, Akshay Krishnamurthy, and Dimitris Papailiopoulos.
\emph{Wait, Wait, Wait\ldots{} Why Do Reasoning Models Loop?} ICML,
2026. arXiv:2512.12895. \url{https://arxiv.org/abs/2512.12895}

{[}15{]} Andy Arditi, Oscar Obeso, Aaquib Syed, Daniel Paleka, Nina
Panickssery, Wes Gurnee, and Neel Nanda. \emph{Refusal in Language
Models Is Mediated by a Single Direction}. arXiv:2406.11717, 2024.
\url{https://arxiv.org/abs/2406.11717}

\end{document}